\documentclass[10pt,journal,compsoc]{IEEEtran}
\usepackage{amsmath,amsfonts}
\usepackage{algorithmic}
\usepackage{algorithm}
\usepackage{array}
\usepackage{textcomp}
\usepackage{stfloats}
\usepackage{url}
\usepackage{verbatim}
\usepackage{graphicx}
\usepackage{cite}

\usepackage{multirow}
\usepackage{makecell}
\usepackage{color}
\usepackage{booktabs}
\usepackage{enumitem}
\usepackage{float}
\usepackage{needspace}
\usepackage{ulem}
\newcommand{\softname}{NVF (ultra)}
\newcommand{\hardname}{NVF (lite)}
\def\eg{\textit{e}.\textit{g}.}
\def\ie{\textit{i}.\textit{e}.}
\def\etal{\textit{et al}.}

\usepackage{epsfig}
\usepackage{amssymb}
\usepackage{amsthm}
\usepackage{pifont}

\usepackage{wrapfig}

\DeclareMathOperator*{\argmin}{arg\,min}
\newtheorem{theorem}{Theorem}[section]
\usepackage{esint}
\usepackage{caption}
\usepackage{subcaption}
\usepackage[dvipsnames]{xcolor}
\usepackage[pagebackref=true,breaklinks=true,colorlinks,bookmarks=false,citecolor=blue]{hyperref}
\usepackage[capitalize]{cleveref}
\crefname{section}{Sec.}{Secs.}
\Crefname{section}{Section}{Sections}
\Crefname{table}{Table}{Tables}
\crefname{table}{Tab.}{Tabs.}

\begin{document}

\title{Neural Vector Fields: Generalizing Distance Vector Fields by Codebooks and Zero-Curl Regularization}

\author{Xianghui Yang, Guosheng Lin~\IEEEmembership{Member,~IEEE,} Zhenghao Chen, Luping Zhou~\IEEEmembership{Senior Member,~IEEE}
\IEEEcompsocitemizethanks{\IEEEcompsocthanksitem Xianghui Yang, Zhenghao Chen and Luping Zhou are with the School of Electrical and Information Engineering, The University of Sydney, Sydney, NSW 2006,
Australia (e-mail: luping.zhou@sydney.edu.au).\protect\
\IEEEcompsocthanksitem Guosheng Lin is with the School of Computer Science and Engineering,
Nanyang Technological University, Singapore 639798.}}
\markboth{Journal of \LaTeX\ Class Files,~Vol.~14, No.~8, August~2015}%
{Shell \MakeLowercase{\textit{et al.}}: Bare Demo of IEEEtran.cls for Computer Society Journals}

\maketitle


\begin{abstract}
Recent neural networks based surface reconstruction can be roughly divided into two categories, one warping templates explicitly and the other representing 3D surfaces implicitly. To enjoy the advantages of both, we propose a novel 3D representation, Neural Vector Fields (NVF), which adopts the explicit learning process to manipulate meshes and implicit unsigned distance function (UDF) representation to break the barriers in resolution and topology. This is achieved by directly predicting the displacements from surface queries and modeling shapes as Vector Fields, rather than relying on network differentiation to obtain direction fields as most existing UDF-based methods do. In this way, our approach is capable of encoding both the distance and the direction fields so that the calculation of direction fields is differentiation-free, circumventing the non-trivial surface extraction step. Furthermore, building upon NVFs, we propose to incorporate two types of shape codebooks, \ie, NVFs (Lite or Ultra), to promote cross-category reconstruction through encoding cross-object priors. Moreover, we propose a new regularization based on analyzing the zero-curl property of NVFs, and implement this through the fully differentiable framework of our NVF (ultra). We evaluate both NVFs on four surface reconstruction scenarios, including watertight vs non-watertight shapes, category-agnostic reconstruction vs category-unseen reconstruction, category-specific, and cross-domain reconstruction.

\end{abstract}

\begin{IEEEkeywords}
Surface Reconstruction, Point Clouds, Mesh, Generalization, Codebook, Curl.
\end{IEEEkeywords}

\section{Introduction}
\label{sec:intro}

\begin{figure}[t]
  \centering
   \includegraphics[width=0.99\linewidth]{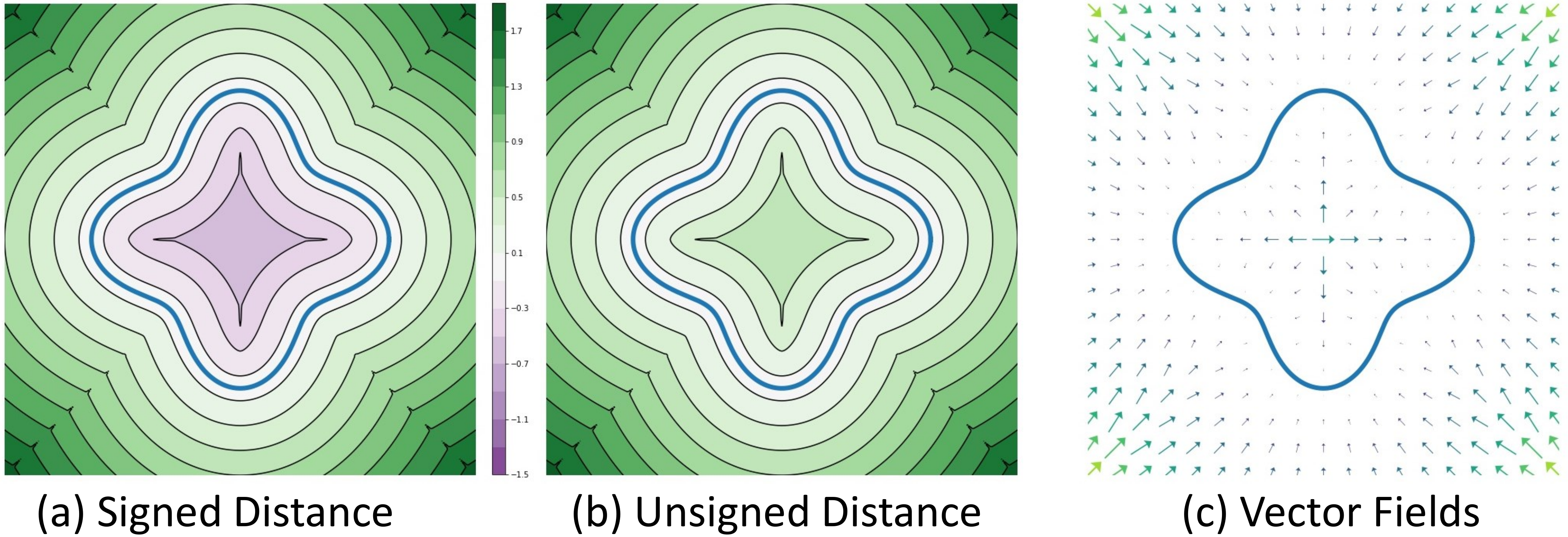}
   \caption{Implicit surface representation methods for watertight shapes: (a) signed distance functions, (b) unsigned distance functions, and (c) our proposed approach vector fields. The blue line indicates the depicted shape.}
   \label{fig:head}
\end{figure}

\IEEEPARstart{R}{econstructing} continuous surfaces is a challenging yet emerging task in various machine-vision applications~\cite{marching_cube, poisson, bspline,bpa} as unstructured, discrete, and sparse point clouds are hard to be deployed into the downstream applications. The remarkable success of Deep Neural Networks (DNNs) has led to the rapid development of several learning-based surface reconstruction methods. These methods can be categorized into \textbf{explicit} or \textbf{implicit} representation methods.

For explicit representation methods, they warp templates~\cite{atlasnet,atlasnet++,nmf,dsp,dgp,Meshlet} or predict voxel grids~\cite{octree, hierarchical, scalable, SkeletonNet} to provide explicit representations, which are friendly to human perception and downstream applications (\eg, editing and rendering). However, these explicit methods are usually limited by topology and resolution. 

On the other hand, implicit representation methods have recently garnered more attention due to their ability to represent surfaces with arbitrary topology, in contrast to explicit methods. Early methods, \cite{points2surf, occnet, ifnet, conv_occnet, local_implicit_grid, learning_implicit_fields, AnalyticMarching} predict the query point as occupancy values or signed distance and then represent the surface as an isocontour of a scalar function as shown in~\cref{fig:head}\textcolor{red}{a}, while generally require specific preprocessing to close non-watertight meshes and remove inner structures.
Recently, Unsigned Distance Functions (UDFs)~\cite{ndf,gifs} are introduced to overcome the high computational preprocessing that eliminates internal structures in the early methods using signed distance functions (SDFs), as shown in~\cref{fig:head}\textcolor{red}{b}. Despite certain advantages, UDFs require a more complicated surface extraction process (Ball-Pivoting Algorithm~\cite{bpa} or gradient-based
Marching Cube~\cite{meshudf, CAP_UDF}) which relies on model differentiation during inference (\ie, differentiation-dependent).

In this work, we propose Neural Vector Fields (NVFs), encompassing the "lite" and "ultra" versions, to leverage the advantages of both the direct manipulation processes from the explicit learning strategy and the powerful representation ability for complex continuous surfaces from the implicit functions for 3D surface reconstruction. Our NVF model the 3D shapes as vector fields and learns a function measuring the displacements of query points to their nearest-neighbor points on the 3D surface. The displacement output of the function could be used directly to deform the source meshes. Therefore, NVFs could serve both as an implicit function and an explicit deformation. Our NVF not only benefits from the advantages of both explicit and implicit representations but also overcomes their drawbacks. 
\underline{Firstly}, NVF representation avoids the comprehensive inference process by skipping the gradient calculation during the surface extraction procedure
\footnote{\label{gradient}. Learning-based methods calculate the gradients of distance fields through model differentiation. The opposite direction of the gradients should point to the nearest-neighbor point on the target surface.} through directly
learning displacement. 
Such a one-pass forward-propagation nature frees NVF from differentiation dependency and significantly reduces the inference time and memory cost. 
\underline{Secondly}, NVFs provide more flexibility to incorporate modules, such as shape codebooks, to promote cross-category surface reconstruction and improve model generalization. Although implicit functions have certain advantages, they often lack the ability to generalize due to their tendency to overfit. We therefore propose to encode cross-category shape priors in two types of shape codebooks. This first one is a hard codebook, dubbed \hardname{}, which consists of un-differentiable discrete shape codes in the embedded feature space using the Vector Quantization (VQ) strategy. As it cuts off gradients, its incorporation with conventional UDFs is retarded as the latter requires gradient information for surface extraction. In contrast, the hard codebook could be seamlessly integrated with NVFs representation thanks to its differentiation-free nature. Such an integration not only improves model generalization but also accelerates the training procedure as a regularization term.
Moreover, we further propose a ``soft" codebook with the cross-attention mechanism, dubbed \softname{}. Compared with the ``hard" codebook that relies on the non-differentiable operation of the nearest search for encoding, our \softname{} leverages the multi-head cross-attention mechanism to combine code items for encoding. This significantly enlarges the space of representation, as well as the perception fields by considering all possible code items in the codebook at the cost of memory. Meanwhile, our soft codebook enables a fully differentiable encoding process, allowing us to jointly optimize the framework and other potential constraints in an end-to-end manner to improve the reconstruction performance.

On top of the end-to-end design in \softname{}, we could further apply zero-curl theory and direction loss as two additional constraints to jointly supervise the point offsets to the surface. It is noted that both the gradient fields of SDF, UDF and the distance vector field of NVF are conservative vector fields and should theoretically satisfy the property of zero curl.  Therefore, we propose a curl regularization to avoid circular vector fields that will result in inaccurate gradients/displacements, leaving uneven surfaces and little holes after surface extraction if the circulations happen near the surface. That is, the divergence of the field should be negative and the curl of the field should be zero (\ie, $\mathbf{\nabla\cdot v}<0$ \& $\mathbf{\nabla\times v}=\mathbf{0}$, $\mathbf{v}$ denotes any vector within the fields). 
To implement curl regularization, we obtain the Jacobian matrix by respectively calculating the first-order partial derivatives of the predicted displacements along x, y, and z directions, followed by computing the curl of distance vector fields and minimizing it to the zero-curl fields. In addition to zero-curl regularisation, we also propose a direction loss to maximize the cosine similarity between the predicted and ground-truth displacements to further surface smoothness.

We conducted extensive experiments on two benchmark reconstruction datasets, which are the synthetic dataset ShapeNet~\cite{shapenet} and the real scanned dataset MGN~\cite{mgn}. We perform our algorithm with other methods in four tasks: category-specific, category-agnostic, category-unseen, and cross-domain reconstruction tasks. The results demonstrate that our~\softname{} framework achieves state-of-the-art reconstruction results in nearly all tasks and all metrics.

Our contributions are summarized as follows.
\begin{itemize}

\item {We propose \textit{NVF} for better 3D field representation, which bridges the explicit learning and implicit representations, and benefits from both of their advantages. Our method can obtain the displacement of a query location in a differentiation-free way, and thus it significantly reduces the inference complexity and provides more flexibility in designing network structures which may include non-differentiable components.}

\item {On top of NVF, we propose to learn shape codebooks, either ``hard'' or ``soft'' to encode shape priors to improve model generalization and cross-category reconstruction. The ``hard'' codebook encodes each query location as a composition of discrete codes in feature space, while the soft one incorporates all coded items via a cross-attention mechanism. The former is computationally less complex, while the latter, although at the cost of increased use of memory, has more flexibility of representation and allows end-to-end optimization with a fully differentiable encoding process.
}
    
\item {We propose two new constraints: curl regularization and direction loss to improve surface quality and prevent large holes by calculating the Jacobian matrix and enforcing zero-curl fields. 
}

\item {We conducted extensive experiments to evaluate the effectiveness of our proposed method. The results demonstrate consistent and promising performance on two benchmarks across various evaluation scenarios. 
}

\end{itemize}

Our preliminary work has been published in a conference paper~\cite{nvf}. This manuscript is a substantial extension of our previous investigation in the following ways. First, we extend the hard codebook used in~\cite{nvf} to a soft codebook by
incorporating all coded items through a cross-attention mechanism. This not only boosts model generalization with more flexibility of encoding shape priors, but also allows end-to-end optimization with a fully differentiable encoding process. Second, this extended study incorporates a comprehensive analysis of the zero-curl and negative divergence properties of the distance vector fields, offering additional regularizations (curl regularization
and direction loss) for field learning. Third, we redo all the experiments on different reconstruction scenarios for our newly proposed soft codebook and regularizations. The culmination of our preliminary research and these novel extensions establish a robust foundation for the advancement of surface reconstruction and ultimately yield state-of-the-art results in surface reconstruction with improved generalization.


\section{Related Work}

\subsection{Explicit Representations}
Voxels~\cite{voxNet,pix3d,3dr2n2} served as the early shape representation method, which can be discretized into grids and processed by subsequent deep learning processing techniques~\cite{surfacenet,stereo_machine,UnsupervisedLO,LearningAP,marrnet}. However, memory footprint usually makes such methods computationally expensive for high-resolution applications and it is non-trivial to reduce such computational costs~\cite{octree,hierarchical,scalable,octnet,volume_mapping}

In contrast, point cloud approaches offer a more efficient means of representing surfaces. Recent learning-based point cloud processing methods~\cite{pointnet,pointnet++,kpconv,point_transformer,edge_conv} have achieved remarkable success in 3D shape analysis~\cite{deep_sets,pointwise,Relation-Shape,RandLA,Deep_Parametric,Minkowski,VoxelNet} and synthesis~\cite{pc_generation,FoldingNet,atlasnet,PCN}. Nevertheless, point clouds often fail to provide detailed geometric information, thereby limiting their usefulness in downstream applications.

In order to incorporate rich geometry information (\eg, surfaces and topology), 3D shapes can be represented as polygonal meshes~\cite{pixel2mesh,GenMesh,learning_category,photometric,Mesh_Autoencoders}. However, traditional polygonal meshes are often characterized by fixed topology, as they deform a predefined template such as sphere~\cite{tmn,learning_category,GenMesh}) into desired shapes by manipulating vertices, which requires non-trivial processes~\cite{atlasnet,meshrcnn,tmn} to alleviate.

In general, distance vector fields offer a distinct advantage over explicit methods, as they rely on explicit deformations using vector fields for learning purposes. This approach effectively circumvents the limitations imposed by the fixed topology and resolution of the templates, allowing greater flexibility in the modeling process.

\subsection{Implicit Representations}
Implicit representation 
can overcome  the limitations of limited resolution and fixed mesh topology
by representing continuous surfaces.
These methods typically employ multi-layer perceptions (MLPs) to generate binary occupancy~\cite{occnet, ldif, ifnet, NASA, PIFu, PIFuHD} or signed distance functions (SDFs)~\cite{deepsdf,local_implicit_grid,learning_implicit_fields} based on spatial locations. 
However, they require extensive pre-processing to artificially close the surfaces and obtain watertight meshes, and lack generalization on non-watertight ones.
In order to address the above issues,
Chibane~\etal~\cite{ndf} introduced unsigned distance functions (UDFs) to learn the unsigned distance, while subsequent methods~\cite{anchor,csp,rangeudf,gifs,CAP_UDF,nvf} are also introduced to improve the performance and generalization of UDFs. 
For instance, GIFS~\cite{gifs} utilizes an intersection classification branch and CAP-UDF~\cite{CAP_UDF} optimizes models directly on raw point clouds. 
Although these UDF-based methods can improve the generalization of implicit representation, they require expensive surface extraction processing~\cite{bpa,meshudf,CAP_UDF}.
Recently, Yang~\etal~\cite{nvf} proposed distance vector fields (DVFs) that encode both distance and direction fields, freeing surface extraction from differentiation.
However, it adopts a vector quantization strategy with a ``hard" codebook design to encode cross-object priors, while non-differentiable operation retards the framework to jointly optimize with other constraints (\eg, curl regularization) in an end-to-end manner.
Our proposed method, \softname{}, replaces the ``hard" codebook with a fully-differentiable soft codebook, which not only
enhance model generalization but also enable the framework to jointly optimize with essential regularization terms.

\subsection{Codebook}
The vector quantization (VQ) is introduced to discretize and compress extracted features into compact representations (\ie, codebook), which was first proposed by Oord~\etal in the VQ-VAE~\cite{vq-vae}.
Subsequently, Yu \etal developed the VQ-GAN~\cite{vqgan}, which utilizes codebooks with adversarial learning for generating high-quality images. Since then, numerous methods have adopted this strategy for various applications such as image generation~\cite{Autoregressive_image,vq-vqe2,CogViewMT,VectorQD}, speech synthesis~\cite{Audio-Visual,jukebox}, video processing~\cite{CogVideo,PredictingVW,LongVG,DLFormerDL}, compression~\cite{Soft-to-Hard,Chen2020DifferentiablePQ}, and the realm of 3D fields~\cite{AutoSDF,autoregressive,3DILG,shapeformer}. 
Despite the tremendous success, the major limitation is that it cannot support end-to-end differentiability, as it only replaces the query with the closest code, thereby halting gradient flow. 
Recently, attention mechanisms have been introduced to make use of all codes in the codebook and improve performance for the corresponding tasks~\cite{codedvtr, coordinates_not_lonely,3DShape2VecSet}.
Our work not only introduces such a ``soft" codebook to enhance feature representation but also leverages this differentiable non-local operation to achieve end-to-end differentiability so that we can further incorporate other constraints (\eg, curl regularization) and jointly optimize them in our framework.

\begin{figure*}[h]
  \centering
   \includegraphics[width=0.99\linewidth]{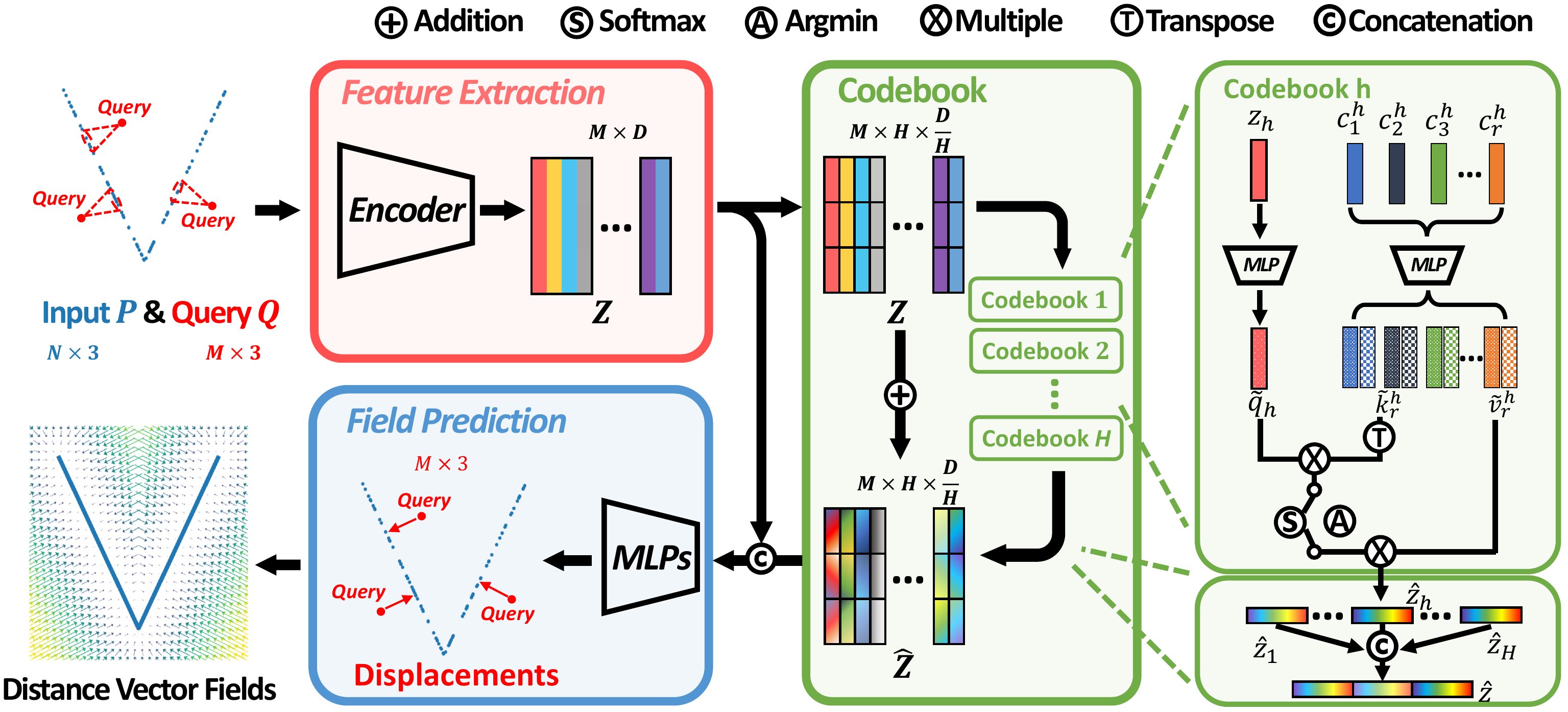}
   \vspace{-1mm}
   \caption{Overview of our NVF (lite and ultra) framework. If the codebook is switched to the~\raisebox{.5pt}{\textcircled{\raisebox{-.9pt} {S}}} mode, the \softname{} will be activated via a $\mathop{\mathrm{softmax}}$ operation; if the codebook is switched to the~\raisebox{.5pt}{\textcircled{\raisebox{-.9pt} {A}}} mode,  the~\hardname{} mode will be activated through an $\mathop{\mathrm{argmin}}$ operation. Our model encodes the input point cloud (blue dots), samples features for each query point (red dots), and fuses query embeddings via either a hard (\hardname{}) or a soft (\softname{}) codebook to predict the displacements of query location to the target surface (red arrow). }
   \label{fig:main}
\end{figure*}

\section{Methodology}
Our objective is to reconstruct the original surface $\mathcal{S}$ from a sparse point cloud $P\in\mathbb{R}^{N\times3}$ using distance vector fields (DVFs) as a representation. These DVFs predict displacement vectors $\mathbf{v}\in\mathbb{R}^3$ that move the point $\mathbf{q}$ to the closest points $\mathbf{\hat{q}}$ on the underlying surface. Our framework consists of three key components, namely, Feature Extraction, Codebook, and Field Prediction, as depicted in~\cref{fig:main}. In particular, for optimizing Field Prediction, we adopt a curl regularization and direction loss as additional constraints. 
The details of each module are further elaborated below.

\subsection{Distance Vector Fields} 
Our NVF (lite and ultra) both adopt a neural network function, which models the underlying shape $\mathcal{S}$ by predicting the field of displacements $\mathbf{v}$ for each given query point $\mathbf{q} \in Q$ to its nearest point $\mathbf{\hat{q}}\in\mathcal{S}$ on the surface.  
The formulation of this method is given by:
\begin{equation}
    \textrm{NVF}(\mathbf{q})=\mathbf{v}={\min_{\hat{q}\in\mathcal{S}}}\ \mathbf{\hat{q}}-\mathbf{q}.
\end{equation}
In general, we produce vector fields as directional distance fields as in~\cite{nvf}, which encode both distance and directional fields $d=||\mathbf{\hat{q}}-\mathbf{q}||_2$ and $\mathbf{g}=(\mathbf{\hat{q}}-\mathbf{q})/||\mathbf{\hat{q}}-\mathbf{q}||_2$.

\subsection{Feature Extraction}\label{sec:FeatureExtraction}
\label{shape_encoding}
Though many existing backbones~\cite{point_transformer,pointnet++,pointnet} could be used as the feature encoder, for simplicity, we directly adopt the feature extraction module in NVF~\cite{nvf} to extract 3D shape features $f_i$ for each point $\mathbf{p}_i$ from the point cloud $P$ and produce the feature embedding $z$ for a given query point $\mathbf{q}$ by concatenating $K$ signatures $z_i$ ($i=1, \cdots, K$), that is, $z=Concat(z_{1},z_{2},...,z_{i},...,z_{K})$. Each signature $z_{i}$ is obtained by nonlinearly transforming a quadruple consisting of the 3D position of $\mathbf{q}$, the 3D positions of $\mathbf{q}$'s $K$ nearest points $\mathbf{p}_i$ on the point cloud $P$, the displacement of $\mathbf{q}$ to $\mathbf{p}_i$, and the feature $f_i$ of $\mathbf{p}_i$. Mathematically, the signature $z_{i}$ can be represented by $z_{i}=MLP(\mathbf{q},\mathbf{p}_i,\mathbf{p}_i-\mathbf{q},f_i), i=1,2,...,K$.


\begin{figure}[h]
  \centering
   \includegraphics[width=0.9\linewidth]{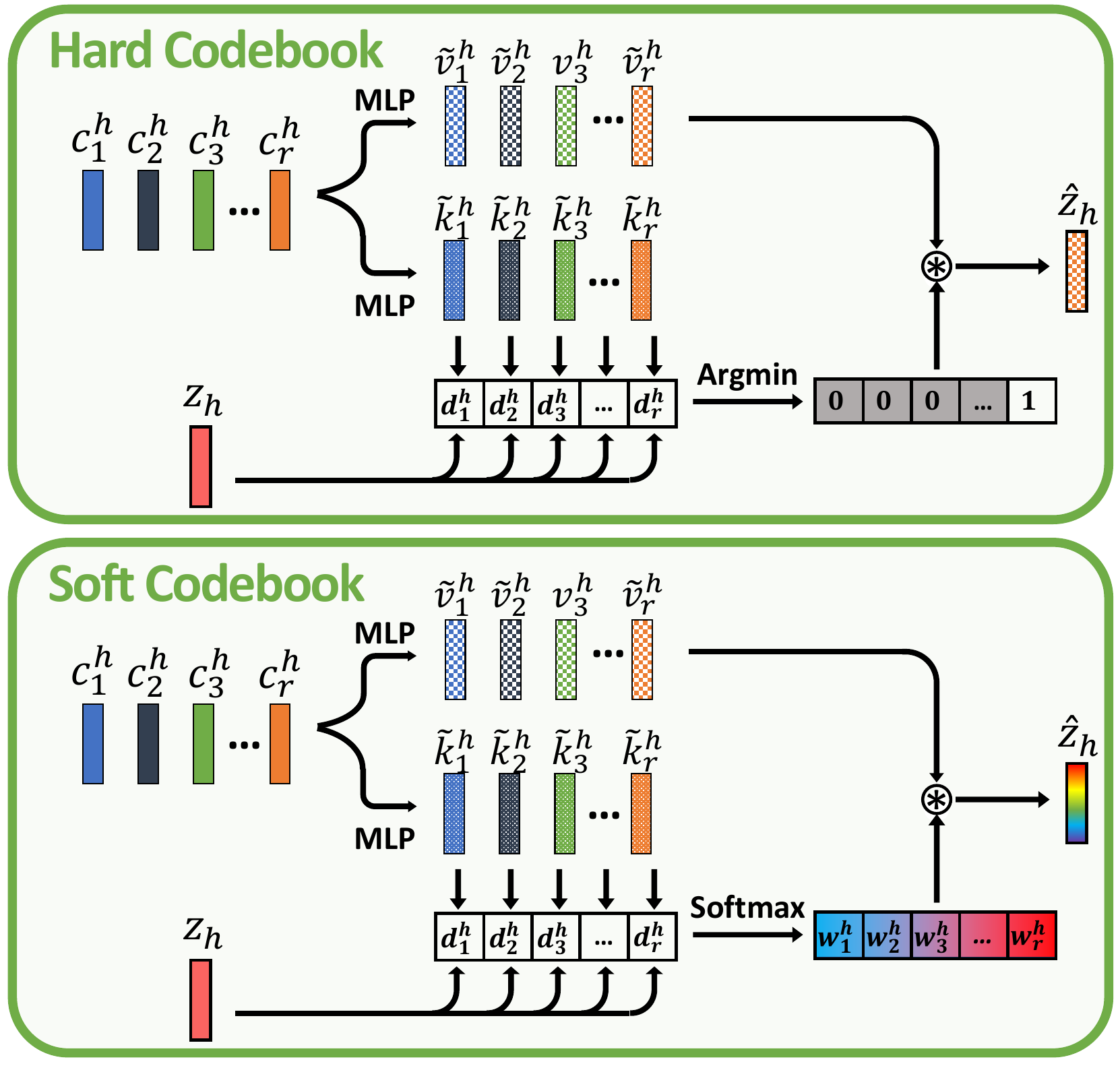}
   \vspace{-1mm}
   \caption{Overview of our proposed hard and soft codebook mechanisms. The hard codebook updates the feature $z_h$ by sampling its nearest code with an $\mathop{\mathrm{argmin}}$ operation, whereas the soft codebook updates the feature $z_h$ by combining all codes based on learned similarity values with a $\mathop{\mathrm{softmax}}$ function.}
   \label{fig:hard_vs_soft}
\end{figure}

\subsection{Codebook} 
A multi-head codebook $\mathcal{C}\in\mathbb{R}^{H\times R\times\frac{D}{H}}$ is learned to extract cross-object priors, while the ``soft'' codebook enhances the representation ability of the ``hard'' codebook used in NVF by using a larger search space for encoding. Here $D$, $H$ and $R$ represent the dimension of the embedding $z$, the number of the sub-codebooks (\ie, heads), and the number of discrete codes $c_r$ ($r=1, \cdots, R$) within each sub-codebook. The dimension of each discrete code is $\frac{D}{H}$.

\textbf{Hard Codebook.} We divide the continuous embedding $z_q$ into $H$ segments along the channel dimension. Each segment is denoted as $z_q^h$. To obtain discrete representations, we find the nearest code in the corresponding sub-codebook ${\mathcal C}^h$ for each segment $z_q^h$, using the Euclidean distance as a measure. This discretization process results in the creation of $\hat{z}_q$, which is composed of discrete segment codes that represent the original continuous embedding $z_q$.
This process can be formulated as follows,
\begin{equation}
\begin{aligned}
 &z_q=z_q^{1}\oplus z_q^{2}\oplus...\oplus z_q^h...\oplus z_q^H,\\
    &\hat{z}_q=c^{1}_{r_1^*}\oplus c^{2}_{r_2^*}\oplus...\oplus c^{h}_{r_h^*}...\oplus c^{H}_{r_H^*}, \\
 & r_h^*=\argmin_{r\in\{1,...,R\}}||z_q^{h}-c^h_r||_2.
\end{aligned}
\end{equation}
When using a single codebook $\mathcal{C}\in\mathbb{R}^{(HR)\times D}$, the feature space is limited in size and lacks diversity. Differently, by adopting a multi-head codebook, $R^H$ permutations are introduced, effectively expanding the feature space from $\mathbb{R}^{HR}$ to $\mathbb{R}^{R^H}$. This augmentation significantly enhances the representation capacity.
Meanwhile, it's important to be aware that the nearest search operation at this stage interrupts the gradient flow. Despite this, the multi-head codebook can still be jointly learned during training. This is achieved by minimizing the distance between the continuous embedding $z$ and its discretization $\hat{z}_q$, which can be formulated as follows:
\begin{equation}
\label{commit}
    \mathcal{L}_{code}=\sum_{h\in\{1,...,H\}} ||sg(c^h_{r^*_h})-z_q^h||_2^2+\beta ||sg(z_q^h)-c^h_{r^*_h}||_2^2,
\end{equation}
where the "stop gradient" operation defined by $sg$ and the weight $\beta$ for the commitment loss in the second term. Additionally, it proposes an alternative approach to update the discretized embedding $\hat{z}_q$ and its corresponding codes ($\hat{z}q^h \equiv c^{h}_{r_h^*}$) in the codebook using the \textit{Exponential Moving Average}~\cite{vq-vae}:
\begin{equation}
    c^h_{r^*_h} := \gamma c^h_{r^*_h} + (1-\gamma) z_q^h,
\end{equation}
with $\gamma$ is a value between 0 and 1.

\textbf{Soft Codebook.} To encode the feature embedding $z$ of the query point ${\mathbf q}$ using the multi-head codebook,  we first split the continuous embedding $z$ 
into $H$ segments along the channel dimension, corresponding to the $H$ sub-codebooks. Each segment of the embedding, denoted as $z^h$ ($h=1, \cdots, H$), is then transformed by MLP into a vector known as \textit{Query}~$\Tilde{q}_h$, while at the same time, each discrete code $c^h_r$ ($r=1, \cdots, R$) in the corresponding $h$-th sub-codebook is also transformed by MLP to form the vectors \textit{Key}~$\Tilde{k}^h_r$ and \textit{Value}~$\Tilde{v}^h_r$, respectively. The three vectors $\Tilde{q}^h$, $\Tilde{k}_r^h$, and $\Tilde{v}_r^h$ are then used to encode $z^h$ by $\hat{z}^h$ through their cross-attention defined in~\cref{eq:Eqn-soft-codebook}.  
\begin{equation}\label{eq:Eqn-soft-codebook}
\begin{aligned}
 &z=z^{1}\oplus z^{2}\oplus...\oplus z^h...\oplus z^H,\\
    &\hat{z}=\hat{z}^1\oplus \hat{z}^2\oplus...\oplus \hat{z}^h...\oplus \hat{z}^H, \\
    &\Tilde{q}^h, \Tilde{k}_r^h, \Tilde{v}_{r}^h=MLP_{q}(z^h), MLP_{k}(c_{r}^h), MLP_{v}(c_{r}^h)\\
 &\hat{z}^h=\sum_{r=1}^{R}softmax(\frac{\Tilde{q}^h \Tilde{k}_r^h}{\sqrt{d_k}})\Tilde{v}_{r}^h
 \end{aligned}
\end{equation}
Here, $d_k=D/H$ is the code dimension and the symbol $\oplus$ denotes concatenation. As can be seen, now $\hat{z}^h$ is encoded by the linear combination of $\Tilde{v}^h_r$ which is transformed from the codes $c_r^h$ in the codebook. Compared with the discrete encoding in the vanilla codebook used by NVF, $\hat{z}^h$ is now a continuous variable, which significantly increases the space of representation and
therefore, further enhances the generalization capacity. Meanwhile, the fully-differentiable cross-attention operation enables our framework to train in an end-to-end manner. The codes stored in the codebook are learnable parameters that can be jointly learned by optimizing final losses in~\cref{sec:optimization}, different from NVF which optimizes an additional loss function to learn the codebook.

\subsection{Field Prediction}
We take the continuous embedding $z$ and its discretization $\hat{z}$ to predict a vector as the displacement for the query point $\mathbf{q}$ to move to the ground truth surface $\mathcal{S}$. By combining the embeddings $z$ and $\hat{z}$, the final vector field is predicted using MLPs,~\ie, $\textrm{NVF}(\mathbf{q})=MLPs(z,\hat{z})$.

\subsection{Curl and Divergence}
In theoretical terms, distance vector fields are required to satisfy the active and irrotational conditions, implying that the divergence of the field should be negative ($\mathbf{\nabla\cdot v<0}$), and the curl of the field should equal zero ($\mathbf{\nabla\times v=0}$). Thus we employ these two operators on the DVFs to depict their active and rotational properties and impose curl regulation during the learning process. Before explaining how to use the curl regularization, we need to establish why the DVFs conform to these conditions. In vector calculus, the curl is a vector operator that characterizes the infinitesimal circulation of a vector field in three-dimensional Euclidean space, while the divergence is a vector operator that performs on a vector field, creating a scalar field indicating the quantity of the vector field's source at each point. These operators describe the outward flux and circulation density of vector fields. We give the following theorem.

\begin{theorem}
\label{theorem:curl}
Given a distance vector field $\mathbf{v}(x,y,z)$ such that $\mathbf{v}(x,y,z)= -\phi\nabla\phi$,  where $\phi(\cdot)$ denotes an unsigned distance field whose gradient is $\nabla\phi$, it can be shown that the distance vector field $\mathbf{v}(x,y,z)$ is an irrotational field and satisfies zero-curl condition $\mathbf{\nabla\times v=\mathbf{0}}$.
\end{theorem}


\begin{proof} 
For a distance vector field $\mathbf{v}(x,y,z)= -\phi\nabla\phi$, where $\phi(\cdot)$ denotes an unsigned distance field whose gradient is $\nabla\phi$, it can be written as the gradient of the potential function $-\frac{1}{2}\phi^2(x,y,z)$, which is a scalar function. That is, 
\begin{equation}
    \mathbf{v}=\nabla(-\frac{1}{2}\phi^2) = -\phi\nabla\phi.
\end{equation}
Therefore, the distance vector field $\mathbf{v}(x,y,z)$ is indeed a \textbf{conservative} vector field. For any \textbf{conservative} vector field, it is an irrotational vector field and thus $\mathbf{v}(x,y,z)$ eventually forms a zero-curl field.
\end{proof}

Intuitively, the magnitude and the direction of the vector field $\mathbf{v}$ are determined by the unsigned distance function $\phi(x,y,z)$ and its gradient $\nabla\phi(x,y,z)$, respectively. It is mathematically formulated as $\mathbf{v}(x,y,z)= -\phi\nabla\phi$. According to Theorem~\ref{theorem:curl}, it is an irrotational vector field and thus should form a zero-curl field.

\begin{figure}[t]
    \centering
  \begin{subfigure}{0.22\textwidth}
    \includegraphics[width=1\textwidth]{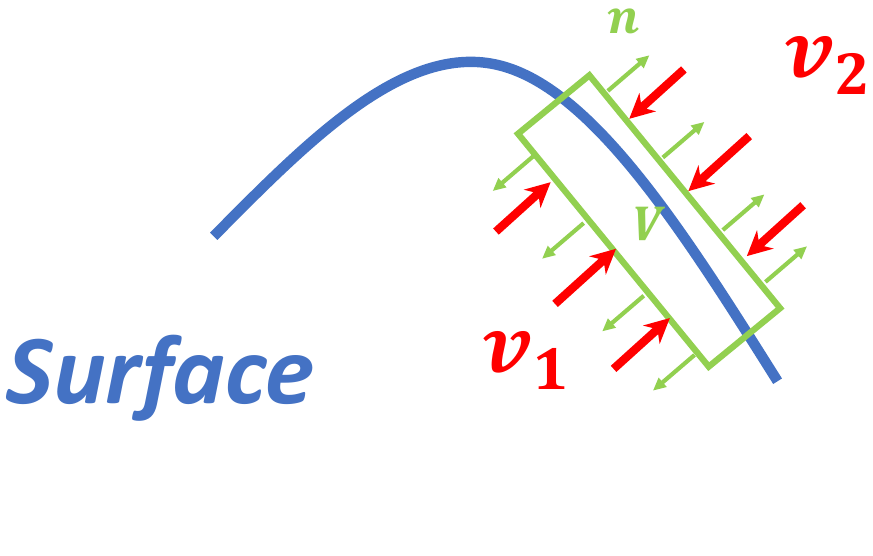}
    \caption{}
    \label{fig:divergence-a}
  \end{subfigure}
  \begin{subfigure}{0.22\textwidth}
    \includegraphics[width=1\textwidth]{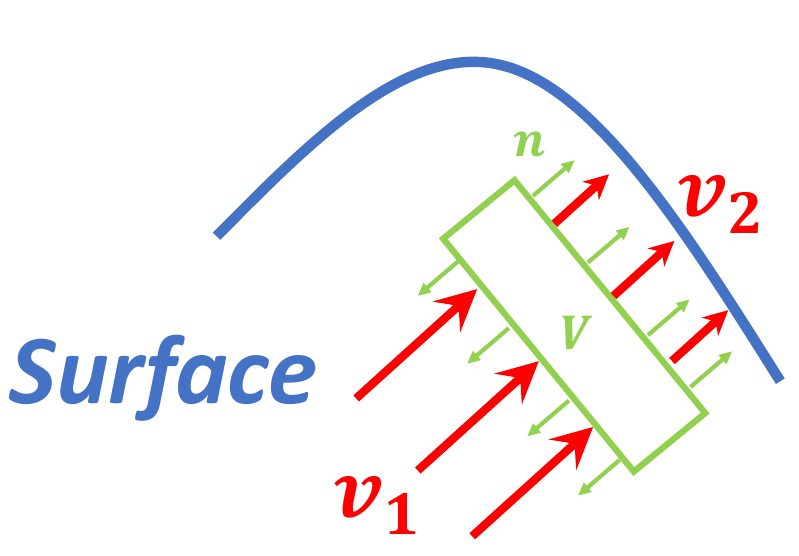}
    \caption{}
    \label{fig:divergence-b}
  \end{subfigure}
  \vspace{-2mm}
   \caption{Illustration of negative divergence condition in a distance vector field. (a) The volume includes the underlying surface, and the vectors are opposite to the surface normals of the volume. (b) The volume does not include the underlying surface, and the magnitude of vectors opposite to surface normals is larger than vectors oriented along with surface normals of the volume. These two scenarios both correspond to negative divergence (~\ie, $\mathbf{\nabla\cdot v<0}$).}
   \label{fig:divergence}
\end{figure}

\begin{figure}[t]
    \centering
  \includegraphics[width=0.45\textwidth]{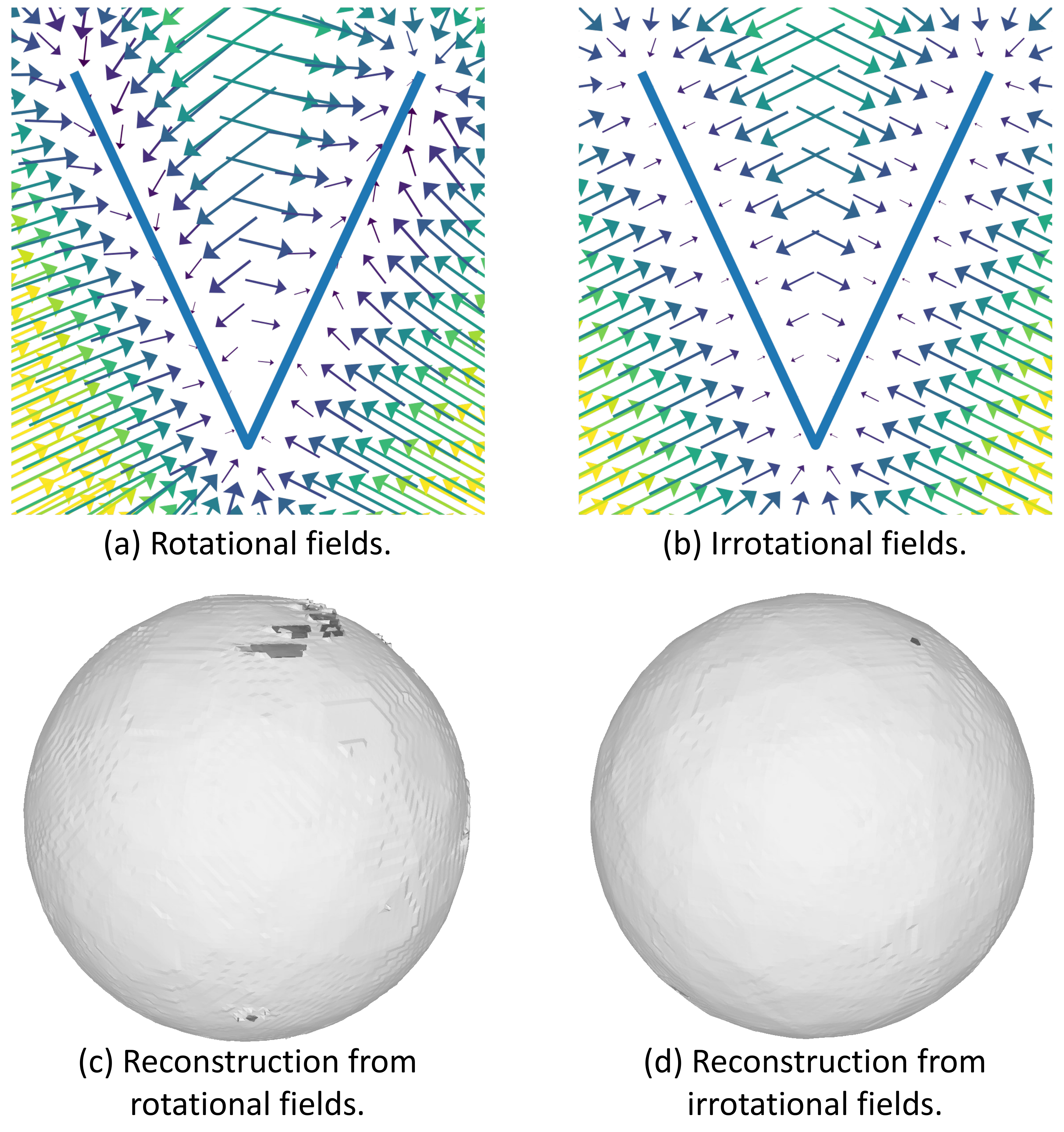}
   \caption{(a) Rotational distance vector fields. (b) Irrotational distance vector fields with zero curl (~\ie, $\mathbf{\nabla\times v=0}$). Reconstruction examples from rotational fields and irrotational fields, where the DVF fits the scene containing a simple sphere (c) with curl or (d) w/o curl regularization. The comparison between (c) and (d) demonstrates that the inaccurate direction of vectors would mislead the surface reconstruction (\eg, leaving holes, and uneven surfaces) as the surface extraction process from distance vector fields relies on the gradient-based marching cubing operation.}
   \label{fig:rotation}
\end{figure}

\begin{theorem}
\label{theorem:divergence}
Given a distance vector field $\mathbf{v}(x,y,z)= -\phi\nabla\phi$, it can be shown that $\mathbf{v}(x,y,z)$ is an active field and satisfies the negative divergence condition $\mathbf{\nabla\cdot v}<0$.
\end{theorem}
\begin{proof}
The divergence of a distance vector field $\mathbf{v}(x,y,z)= -\phi\nabla\phi$ at any 3D point $\mathbf{x}$ is defined as the ratio limit of the surface integral of $\mathbf{v(x)}$ out of the closed surface of volume $V$, which encloses the given point $\mathbf{x}$, as $V$ shrinks to zero. This can be formulated as,
\begin{equation}
    \mathbf{\nabla\cdot v}=\lim_{V\to 0}\frac{1}{|V|}\varoiint_{S(V)}\mathbf{v}\cdot\mathbf{n}dS.
    \label{eq:divergence}
\end{equation}
$|V|$ and $S(V)$ are the volume and the boundary of $V$. $\mathbf{n}$ is the outward unit normal to that surface of $V$.

Assuming the volume includes the underlying surface as shown in~\cref{fig:divergence-a}, the distance vectors at the boundary $S(V)$ point inward, in opposition to the surface normals of the closed volume, which yields $\mathbf{v}\cdot\mathbf{n}<0$ and results in the integral in~\cref{eq:divergence} negative. 

In contrast, assuming the volume does not include the underlying surface as shown in~\cref{fig:divergence-b}, the magnitude of vectors on the closer side to the target shape (\ie, $\mathbf{v_2}$ in~\cref{fig:divergence-b}) is less than that of on the far side (\ie, $\mathbf{v_1}$ in~\cref{fig:divergence-b}), which implies $||\mathbf{v}_2||_2<||\mathbf{v}_1||_2$. Since $\mathbf{v_1}\cdot\mathbf{n}<0$ and $\mathbf{v_2}\cdot\mathbf{n}>0$, we will have $\mathbf{v_1}\cdot\mathbf{n} + \mathbf{v_2}\cdot\mathbf{n}<0$. Therefore, the integral in~\cref{eq:divergence} is also negative. 
\end{proof}

Theorem~\ref{theorem:divergence} shows that the divergence of the distance vector field $\mathbf{v}(x,y,z)$ is negative (\ie, $\nabla\times\mathbf{v}<0$). Note that, although we show $\mathbf{v}(x,y,z)$ is an active field, we did not use the negative divergence condition to regularize our reconstruction due to its complexity for optimization. This could be further explored in our future work.

In addition, we present an example to visualize the rotational and irrotational fields in~\cref{fig:rotation}\textcolor{red}{(a,b)}, with particular emphasis on their behavior near the object. 
For better comparison, we employ a 4-layer MLP model to fit a sphere under L1 supervision to facilitate an equitable comparison of the reconstructed results from both using and without using curl regularization, as depicted in~\cref{fig:rotation}\textcolor{red}{(c,d)}. The reconstructed surfaces affirm the efficacy of incorporating curl regularization for achieving smoothness and bridging small gaps in the resulting surfaces.


It is important to note that the active property has already been taken into account in the existing learning process, as the primary objective is to fit a distance field, and the distance loss can serve as divergence regularization. However, to ensure that the model conforms to the irrotational property, additional measures need to be taken. The properties mentioned above are present exclusively in differentiable regions since they are also non-differentiable areas. A non-curl DVF is shown in~\cref{fig:rotation}. It has been established that distance vector fields satisfying the zero-curl and negative divergence conditions are ideal, and it is desirable to incorporate these properties into the optimization process.

Assuming the predicted offsets composed of elements on three axes are $\mathbf{v}=(v_x, v_y, v_z)$, where $v_x, v_y, v_z$ are functions of position $(x,y,z)$. The curl operator on the distance vector fields can be formulated as,
\begin{equation}
\begin{aligned}
    &\mathbf{\nabla\times v}=\left[\begin{array}{ccc}
     \mathbf{i} & \mathbf{j} & \mathbf{k}\\ 
     \frac{\partial}{\partial x} & \frac{\partial}{\partial y} & \frac{\partial}{\partial z}\\
     v_x & v_y & v_z
    \end{array}\right]\\
    &=(\frac{\partial v_z}{\partial y}-\frac{\partial v_y}{\partial z})\mathbf{i}+ (\frac{\partial v_x}{\partial z}-\frac{\partial v_z}{\partial x})\mathbf{j}+(\frac{\partial v_y}{\partial x}-\frac{\partial v_x}{\partial y})\mathbf{k}\\
    &=\mathbf{0}
\end{aligned}
\end{equation}
 where $\mathbf{i,j,k}$ are unit vectors along the x-, y-, and z-axes, respectively. It is observed that enforcing the curl of the field to be zero is equivalent to forcing their components to be zero, defined as, 
 \begin{equation}
  \frac{\partial v_z}{\partial y}=\frac{\partial v_y}{\partial z}, \frac{\partial v_x}{\partial z}=\frac{\partial v_z}{\partial x}, \frac{\partial v_y}{\partial x}=\frac{\partial v_x}{\partial y}
  \label{eq:partial}
 \end{equation}
 To obtain the first-order partial derivates in~\cref{eq:partial}, we must calculate the Jacobian matrix which can be defined as, 
 \begin{equation}
\mathbf{J(\mathbf{v})}=\left[\begin{array}{ccc}
\frac{\partial v_x}{\partial x} & \frac{\partial v_x}{\partial y} & \frac{\partial v_x}{\partial z} \\
\frac{\partial v_y}{\partial x} & \frac{\partial v_y}{\partial y} & \frac{\partial v_y}{\partial z} \\
\frac{\partial v_z}{\partial x} & \frac{\partial v_z}{\partial y} & \frac{\partial v_z}{\partial z}
\end{array}\right]
\end{equation}
It is observed that encouraging the curl of the field to be zero is equivariant to encouraging the Jacobian matrix to be a symmetry matrix. The curl regularization is,
\begin{equation}
    \mathcal{L}_{curl} = ||\mathbf{J(\mathbf{v})}-\mathbf{J(\mathbf{v})}^{T}||_F
\end{equation}
where $||\cdot||_F$ denotes Frobenius Norm and it is defined as the square root of the sum of the absolute squares of its elements,~\eg, $||A||_F=\sqrt{\sum_i^{m}\sum_j^{n}||a_{ij}||^2}, A\in\mathcal{R}^{m\times n}$.

\subsection{Optimization.}
\label{sec:optimization}
\textbf{Lite.} 
To optimize \hardname{}, we focus on minimizing both the predicted displacement $\mathbf{\Delta q}$ and the discrepancy with the ground-truth displacement $\mathbf{\hat{q}-q}$. Additionally, we consider the discretization error of the continuous embedding $z_q$ as it transitions into $\hat{z}_q$. In order to prevent gradient flow beyond vector quantization, we utilize the "stop-gradient" operation (\textit{sg}).
The comprehensive objective function can be expressed as follows:
\begin{equation}
    \mathcal{L}=\sum_{\mathbf{q}\in Q} ||\mathbf{\hat{q}}-\mathbf{q}-\mathbf{\Delta q}||_1 + \lambda \sum ||z_q-sg(\hat{z}_q)||_2^2.
\end{equation}
We use a hyper-parameter the parameter $\lambda$ to determine the balance between the two loss terms. The commitment loss term $||sg(z)-z^q||_2^2$ in Equation~\ref{commit} has been replaced with the \textit{Exponential Moving Average} algorithm. This replacement is applied to update the corresponding codes in the multi-head codebook.

\textbf{Ultra} We optimize ~\softname{} by minimizing the difference between the predicted displacement $\mathbf{v}$ and the ground-truth displacement $\mathbf{\hat{q}}-\mathbf{q}$, curl regularization $\mathcal{L}_{curl}$, and the direction loss $\mathcal{L}_{dir}$ measuring the directional error of $\mathbf{v}$ and $\mathbf{\hat{q}}-\mathbf{q}$. In sum, our overall objective function is as follows.
\begin{equation}
    \mathcal{L} =  \mathcal{L}_{l_1} + \lambda_{curl}\mathcal{L}_{curl} +  \lambda_{dir}\mathcal{L}_{dir}
\end{equation}
where $\mathcal{L}_{l_1}=||\mathbf{v}-(\mathbf{\hat{q}}-\mathbf{q})||_1$ and $\mathcal{L}_{dir}=1-\frac{\mathbf{v}\cdot(\mathbf{\hat{q}}-\mathbf{q})}{||\mathbf{v}||_2||\mathbf{\hat{q}}-\mathbf{q}||_2}$. The hyperparameters $\lambda_{curl}$ and $ \lambda_{dir}$ balance the loss magnitude.

\subsection{Surface Extraction}



Both signed and unsigned distance functions define the surfaces as a 0-level set. SDFs extract surfaces through Marching Cubes~\cite{marching_cube} faithfully due to the intersection being easily captured via sign flipping. Recently, MeshUDF~\cite{meshudf} and CAP-UDF~\cite{CAP_UDF} propose to detect the opposite gradient directions to replace the sign in SDFs and successfully apply the Marching Cubes~\cite{marching_cube} on UDFs. Note that these two methods learn distance fields and they need to differentiate the distance field to obtain the gradient direction. In contrast, our NVF can similarly extract surfaces using Marching Cubes while avoiding the differentiation of the distance field. This is because our NVF directly encodes both distance and direction fields. The surface extraction algorithm divides the space into grids and decides whether the adjacent corners are located on the same side or two sides. NVF predicts the vectors $\mathbf{\Delta q}$ of all lattices and normalizes them into normal vectors $\mathbf{g_i}=\Delta \mathbf{q_i}/||\Delta \mathbf{q_i}||_2$. Given a lattice gradient $\mathbf{g_i}$ and its adjacent lattices' gradients $\mathbf{g_j}$, the algorithm checks if their gradients have opposite orientations, and assigns the pseudo sign $s_i$ to the lattice. For more details, please refer to MeshUDF~\cite{meshudf}.

\begin{table*}[t]
\centering
\resizebox{0.99\textwidth}{!}{
    \begin{tabular}{@{}c|c|c|c|c|c|c|c|c|c|c@{}}
    \hline
    \multirow{2}{*}{Methods} & \multicolumn{5}{c|}{Base} & \multicolumn{5}{c}{Novel}\\
    \cline{2-11}
    & \multicolumn{1}{c|}{CD$\downarrow$} & \multicolumn{1}{c|}{EMD$\downarrow$} & \multicolumn{1}{c|}{Normal$\downarrow$} & \multicolumn{1}{c|}{F1$_{2.5\times10^{-5}}\uparrow$} & \multicolumn{1}{c|}{F1$_{1\times10^{-4}}\uparrow$}& \multicolumn{1}{c|}{CD$\downarrow$} & \multicolumn{1}{c|}{EMD$\downarrow$} & \multicolumn{1}{c|}{Normal$\downarrow$} & \multicolumn{1}{c|}{F1$_{1\times10^{-5}}\uparrow$} & \multicolumn{1}{c}{F1$_{2\times10^{-5}}\uparrow$} \\
    \hline
    Input & 8.398 & 1.045 & N/A & 14.148 & 25.111 & 7.999 & 1.024 & N/A & 17.576 & 29.815 \\
    OccNet~\cite{occnet} & 27.658 & 1.694 & 0.237 & 30.877 & 46.644 & 447.620 & 4.013 & 0.455 & 15.943 & 24.433 \\
    ConvOccNet~\cite{conv_occnet} & 5.479 & 1.210 & 0.155 & 51.872 & 71.448 & 17.525 & 1.669 & 0.225 & 44.165 & 61.417 \\
    IF-Net~\cite{ifnet} & 1.897 & 1.120 & 0.098 & 65.975 & 85.421 & 5.961 & 1.608 & 0.137 & 61.670 & 81.106 \\
    NDF~\cite{ndf} & 1.689 & 1.538 & 0.110 & 66.802 & 84.809 & 1.694 & 1.741 & 0.135 & 65.622 & 84.069 \\
    GIFS~\cite{gifs} & 1.793 & 1.280 & 0.161 & 56.188 & 78.458 & 1.943 & 1.534 & 0.177 & 56.644 & 78.016 \\
    \hline
    Ours (lite) & 0.910 & 1.079 & 0.108 & 78.503 & 91.408 & 1.442 & 1.145 & 0.109 & 80.883 & 91.836 \\
     Ours (ultra) & ~\textbf{0.791}& ~\textbf{1.069} & ~\textbf{0.079}  & ~\textbf{80.517} & ~\textbf{92.859} & ~\textbf{1.361} & ~\textbf{1.097} & ~\textbf{0.083} & ~\textbf{82.622} & ~\textbf{93.031}\\

     \hline
    \end{tabular}
    }
    \caption{Quantitative results of category-agnostic and category-unseen reconstructions on watertight shapes of ShapeNet. We train all models on the base classes, and evaluate them on the base and the novel classes, respectively.}
    \label{tab:watertight}
\end{table*}

\begin{table*}[t]
\centering
\resizebox{0.99\textwidth}{!}{
    \begin{tabular}{@{}c|c|c|c|c|c|c|c|c|c|c@{}}
    \hline
    \multirow{2}{*}{Methods} & \multicolumn{5}{c|}{Base} & \multicolumn{5}{c}{Novel}\\
    \cline{2-11}
    & \multicolumn{1}{c|}{CD$\downarrow$} & \multicolumn{1}{c|}{EMD$\downarrow$} & \multicolumn{1}{c|}{Normal$\downarrow$} & \multicolumn{1}{c|}{F1$_{2.5\times10^{-5}}\uparrow$} & \multicolumn{1}{c|}{F1$_{1\times10^{-4}}\uparrow$}& \multicolumn{1}{c|}{CD$\downarrow$} & \multicolumn{1}{c|}{EMD$\downarrow$} & \multicolumn{1}{c|}{Normal$\downarrow$} & \multicolumn{1}{c|}{F1$_{1\times10^{-5}}\uparrow$} & \multicolumn{1}{c}{F1$_{2\times10^{-5}}\uparrow$} \\
    \hline
    Input & 3.170 & 0.867 & N/A & 32.875 & 51.105 & 2.885 & 0.843 & N/A & 39.902 & 58.092 \\
    NDF~\cite{ndf} & 0.986 & 1.372 & 0.143 & 72.425 & 88.754 & 0.926 & 1.532 & 0.171 & 76.162 & 89.977 \\
    GIFS~\cite{gifs} & 1.175 & 1.260 & 0.197 & 64.915 & 85.115 & 2.960 & 1.499 & 0.230 & 69.252 & 86.518 \\
    
     \hline
    Ours (lite) & 0.854 & 1.197 & 0.162 & 75.372 & 90.266 & 0.776 & 1.340 & 0.169 & 79.723 & 91.576\\
     Ours (ultra) & \textbf{0.810} & \textbf{1.182} & \textbf{0.137} & \textbf{76.256} & \textbf{90.785} & \textbf{0.745} & \textbf{1.309} & \textbf{0.149} & \textbf{80.349} & \textbf{91.897}\\
     \hline
    \end{tabular}
    }
    \caption{Quantitative results of category-agnostic and category-unseen reconstructions on non-watertight shapes of ShapeNet. We train all models on the base classes and evaluate them on the base and the novel classes, respectively.}
    \label{tab:non_watertight}
\end{table*}

\begin{table}[t]
\centering
\resizebox{0.48\textwidth}{!}{
    \begin{tabular}{@{}c|c|c|c|c|c@{}}
    \hline
    Methods & CD$\downarrow$ & EMD$\downarrow$ & Normal$\downarrow$ & F1$_{1\times10^{-5}}\uparrow$ & F1$_{2\times10^{-5}}\uparrow$ \\
    \hline
    Input & 3.631 & 0.707 & N/A & 23.735 & 41.588 \\
    NDF~\cite{ndf} & 1.971 & 1.248 & 0.275 & 64.116 & 84.902 \\
    GIFS~\cite{gifs} & 1.461 & 0.970 & 0.302 & 54.867 & 79.722 \\
    \hline
     Ours (lite) & 1.144 & \textbf{0.945} & 0.286 & 64.261 & 85.290 \\
     Ours (ultra) & \textbf{1.082}& 0.977 & ~\textbf{0.244} & \textbf{65.993} & \textbf{86.486} \\
     \hline
    \end{tabular}
    }
    \caption{Quantitative evaluation on ShapeNet Cars. We train and evaluate our method on the raw data of the ShapeNet “Car” category. Our method achieves better performance than the state-of-the-art UDF-based methods. }
    \label{tab:car}
\end{table}

\begin{table}[!h]
\centering
\resizebox{0.49\textwidth}{!}{
    \begin{tabular}{@{}c|c|c|c|c|c@{}}
    \hline
    Methods & CD$\downarrow$ & EMD$\downarrow$ & Normal$\downarrow$ & F1$_{1\times10^{-5}}\uparrow$ & F1$_{2\times10^{-5}}\uparrow$ \\
    \hline
    Input & 1.243 & 0.157 & N/A & 52.189 & 72.969 \\
    NDF~\cite{ndf} & 0.252 & 0.216 & 0.097 & 96.338 & 98.687 \\
    GIFS~\cite{gifs} & 0.394 & 0.192 & 0.128 & 93.330 & 97.295 \\
    
     \hline
    Ours (lite) & 0.141 & 0.184 & 0.083 & 98.499 & 99.498 \\
     Ours (ultra) & \textbf{0.133} & \textbf{0.167} & \textbf{0.070} & \textbf{98.756} & \textbf{99.562}\\
    \hline
    \end{tabular}
    }
    \caption{Quantitative results of cross-domain reconstruction on MGN~\cite{mgn}. We train our models based on ShapeNet with the base classes and evaluate them on the raw data from MGN.}
    \label{tab:garment}
\end{table}

\begin{figure*}[h]
  \centering
  \includegraphics[width=0.95\linewidth]{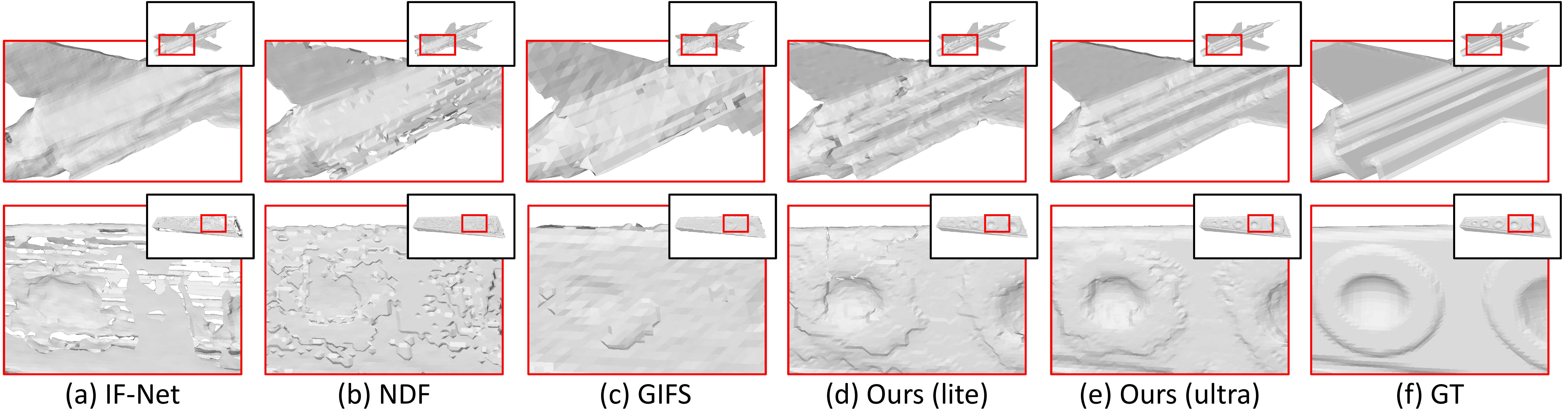}
   \caption{Visualization of category-agnostic and category-unseen reconstructions on watertight shapes from the ShapeNet dataset. The $1^{st}$ row, planes, is from the base classes. The $2^{rd}$ row, speaker, is from the novel classes. We zoom in on part of the shapes (indicated by the red box) to visualize details better.}
   \label{fig:watertight}
\end{figure*}

\begin{figure*}[h]
  \centering
  \includegraphics[width=0.9\linewidth]{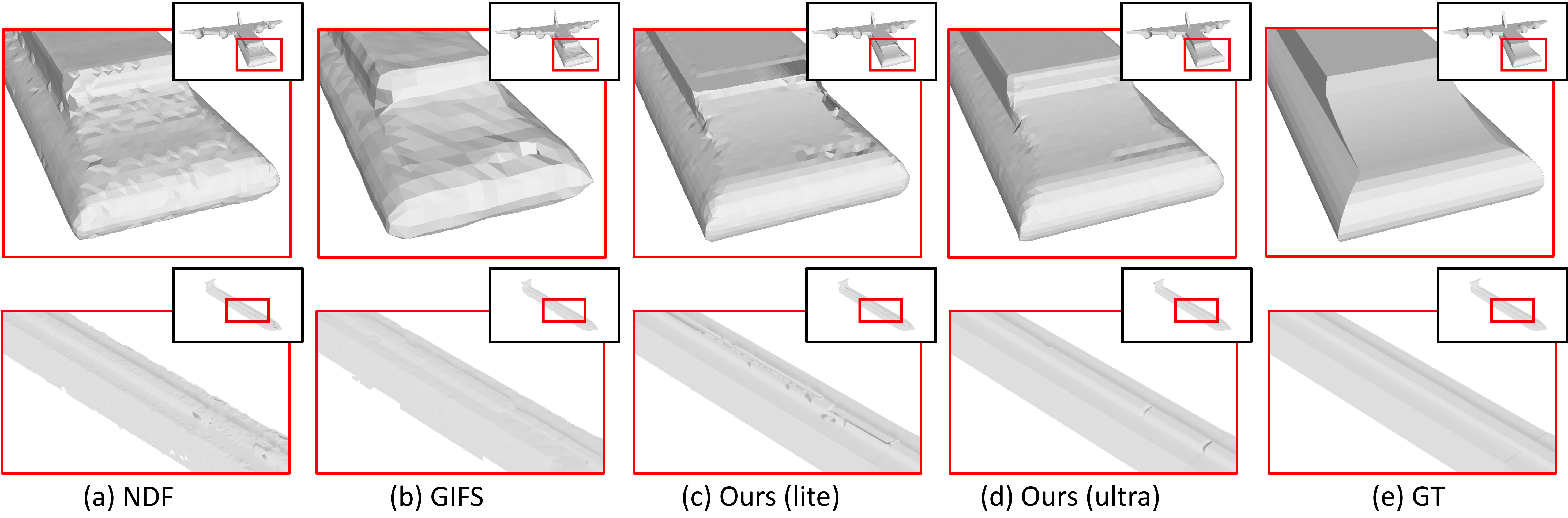}
   \caption{Visualization of category-agnostic and category-unseen reconstructions on non-watertight shapes from the ShapeNet dataset. The $1^{st}$ row, airplanes, is from the base classes. The $2^{nd}$ row, watercraft, are from the novel classes. We zoom in on part of the shapes (indicated by the red box) to visualize details better.}
   \label{fig:non_watertight}
\end{figure*}

\begin{figure}[h]
  \centering
  \includegraphics[width=0.99\linewidth]{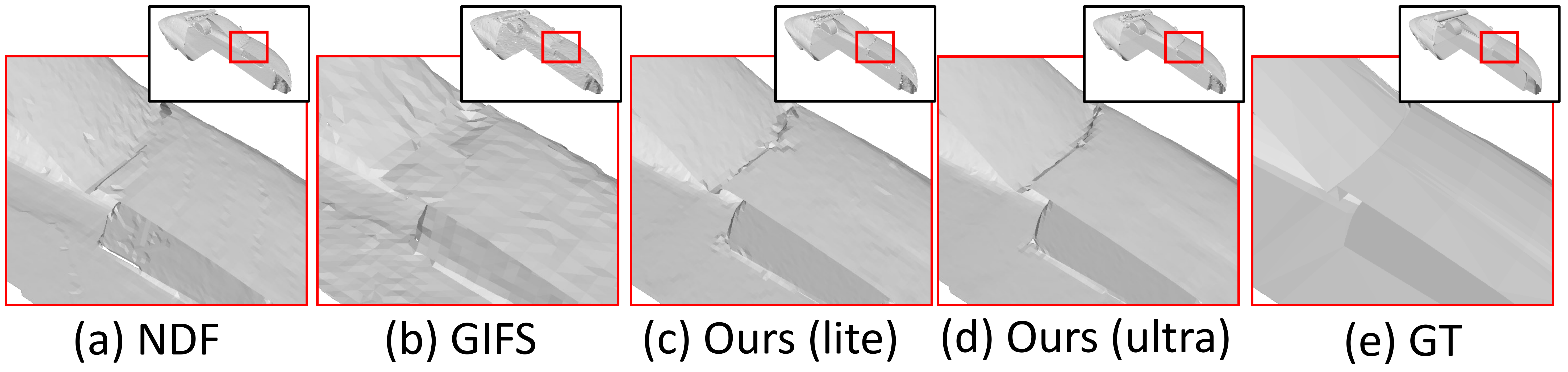}
   \caption{Qualitative visualization of Category-specific reconstruction on ShapeNet Cars. We zoom in on part of the shape (indicated by the red box) to visualize inner structures better.}
   \label{fig:car}
\end{figure}

\begin{figure}[h]
  \centering
  \includegraphics[width=0.99\linewidth]{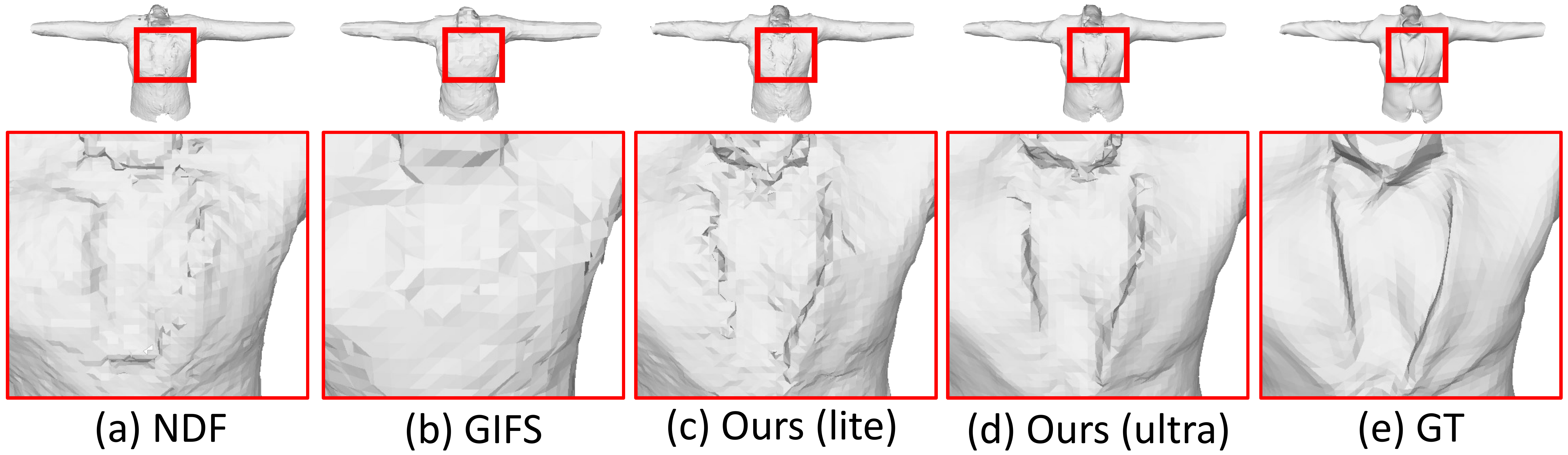}
  \caption{Visualization of cross-domain reconstruction on the MGN dataset. All models are trained on the base classes of ShapeNet without fine-tuning on MGN. We zoom in on part of the shapes (indicated by the red box) to visualize details better.}
   \label{fig:garment}
\end{figure}

\begin{table*}[t]
\centering
\resizebox{0.99\textwidth}{!}{
    \begin{tabular}{@{}l|c|c|c|c|c|c|c|c|c|c@{}}
    \hline
    \multirow{2}{*}{Components}& \multicolumn{5}{c|}{Base} & \multicolumn{5}{c}{Novel}\\
    \cline{2-11}
     &\multicolumn{1}{c|}{CD$\downarrow$} & \multicolumn{1}{c|}{EMD$\downarrow$} & \multicolumn{1}{c|}{Normal$\downarrow$} & \multicolumn{1}{c|}{F1$_{2.5\times10^{-5}}\uparrow$} & \multicolumn{1}{c|}{F1$_{1\times10^{-4}}\uparrow$}& \multicolumn{1}{c|}{CD$\downarrow$} & \multicolumn{1}{c|}{EMD$\downarrow$} & \multicolumn{1}{c|}{Normal$\downarrow$} & \multicolumn{1}{c|}{F1$_{1\times10^{-5}}\uparrow$} & \multicolumn{1}{c}{F1$_{2\times10^{-5}}\uparrow$} \\
    \hline
    Vanilla & 0.121 & 1.212 & 0.196 & 73.642 & 88.962 & 0.081 & 1.329 & 0.193 & 78.800 & 91.084 \\
    \hline
    +Hard & 0.854 & 1.197 & 0.162 & 75.372 & 90.266 & {0.776} & 1.340 & 0.169 & 79.723 & 91.576\\
     +Soft & \textbf{0.803} & {1.197}  & {0.143} & {76.117} & {90.756} & 0.818 & {1.307} & {0.153} & {80.254} & {91.894}\\
     +Soft +Curl & {0.804} & 1.190 & 0.144 & 76.072 & 90.718 & 0.841 & \textbf{1.291} & 0.153 & 80.265 & \textbf{91.898}\\
     +Soft +Curl +Dir. & 0.810 & \textbf{1.182} & \textbf{0.137} & \textbf{76.256} & \textbf{90.785} & \textbf{0.745} & 1.309 & \textbf{0.149} & \textbf{80.349} & 91.897\\
     \hline
    \end{tabular}
    }
    \caption{Quantitative results of alternative NVF (ultra) methods with different proposed components for category-agnostic and category-unseen reconstructions on non-watertight shapes of ShapeNet. We train all models on the base classes and evaluate them on the base and the novel classes, respectively.}
    \label{tab:abalation}
\end{table*}


\section{Experiments}
\subsection{Experimental Protocol}

\noindent\textbf{Lite and Ultra.} When considering the usage of the soft codebook and training with curl and direction loss, it is important to acknowledge the significant increase in memory requirements and time consumption. In light of these factors, we have designated this approach as NVF (ultra). In contrast, the employment of a hard codebook involves sampling the closest code based on Euclidean distance. While this method is not compatible with curl loss, it only necessitates a minimal additional memory allocation. Consequently, we refer to it as NVF (lite).

\noindent\textbf{Tasks.} We evaluate the effectiveness of our framework across four distinct reconstruction tasks:  1) category-agnostic, 2) category-unseen, 3) category-specific,  and 4) cross-domain. In Section \ref{sec:agnostic}, we first assess the generalization ability of our~\hardname{} and ~\softname{} against existing methods through category-agnostic and category-unseen reconstruction. In Section \ref{sec:specific}, we demonstrate the NVF's capacity for reconstructing non-watertight meshes through category-specific reconstruction. Finally, in Section \ref{sec:cross}, we perform a cross-domain reconstruction test by reconstructing real scanned data without any training or fine-tuning.

\noindent\textbf{Implementations.} We follow~\cite{nvf} to use the PointTransformer~\cite{point_transformer} with $K=16$ nearest points as our feature encoder and set the grid resolution to $256$ for surface extraction. For codebook design, 
We respectively set the number of heads as 4 and the number of codes and dimensions in each sub-codebook as 256 and 64.
During the optimization, we set the learning rate to $l_r=10^{-4}$ which decays by a factor of $0.3$ at steps 30, 70, and 120. Then we apply curl regularization from step $150$, with weights $\lambda_{curl}$ as $10^{-6}$ and $\lambda_{dir}$ as $10^{-3}$. The Adam optimizer has been utilized for training with a learning rate of $10^{-3}$, $(\beta_1, \beta_2)=(0.9, 0.999)$, $\epsilon=10^{-8}$, and weight decay $\lambda=0$.

\noindent\textbf{Baseline method implementation.} For a fair comparison, we used the publicly released codes of the baseline methods using identical training/testing conditions and the most optimal hyperparameter sets suggested in corresponding papers. We pre-process the shape as DISN~\cite{DISN} and assign sampled query points inside/outside labels according to the codes released by OccNet~\cite{occnet} and IF-Nets~\cite{ifnet} for optimizing watertight shape reconstruction. We adopt the same 3D convolution network as the backbone for IF-Nets~\cite{ifnet}, NDF~\cite{ndf}, and GIFs~\cite{gifs}. Their input point clouds are all voxelized with a resolution of $128^3$, consistent with the watertight car reconstruction based on~\cite{ndf, gifs}. For non-watertight shape reconstruction, such as category-agnostic, category-unseen, category-specific, and cross-domain reconstructions, the input point clouds of the three aforementioned methods are voxelized with a resolution of $256^3$

\noindent\textbf{Datasets.} 
We conduct our category-agnostic, category-unseen, and category-specific reconstruction experiments on ShapeNet~\cite{shapenet}, a large synthetic dataset, whereas we evaluate our model's generalization ability on MGN~\cite{mgn},  a real scanned dataset, through cross-domain reconstruction experiments.
All meshes are normalized to a unit cube with a range of $[-0.5, 0.5]$, consistent with existing methods.

ShapeNet is a synthetic dataset with 55 distinct object categories. 
We use \textit{cars, chairs, planes, and tables} (referred to as \textit{base classes}) categories for category-agnostic reconstruction. We also perform category-unseen reconstruction on the \textit{speakers, bench, lamps, and watercraft} (referred to as the \textit{novel classes}) categories in~\cref{tab:watertight} and~\cref{tab:non_watertight}. Last, we use the \textit{cars} category to conduct our category-specific reconstruction experiments in~\cref{tab:car}. 
While, the MGN dataset comprises 5 garment categories, with open surfaces obtained through scanning. We evaluate our framework on MGN to demonstrate our model generalization in the wild.

\noindent\textbf{Data Preprocessing.} We generate training point samples,~\ie, query points, along with the corresponding ground-truth displacement vectors. Specifically, we pre-compute 100k points according to the UDF method as in~\cite{ndf}. To capture surface details, we sample points in the vicinity of the surface by randomly selecting a surface point $\hat{\mathbf{q}}$ from the ground-truth mesh, which is displaced by a vector $\mathbf{d}$ drawn from a Gaussian distribution $\mathbf{d}\sim N(0,\Sigma))$. In this way, we can create a 3D query point $\mathbf{q}=\hat{\mathbf{q}}+\mathbf{d}$. We use a diagonal covariance matrix $\Sigma\in\mathbb{R}^{3\times 3}$ with entries $\Sigma_{i,i} = \sigma$. 

During the training stage, we use a subset of 30k samples, where $1\%$ of them generated by $\sigma = 0.08$ for optimizing the displacement predictions for query points far from the surface, $49\%$ generated by $\sigma=0.02$ for optimizing regression of surface distances within {$\sigma=0.02$, and $50\%$ generated by $\sigma = 0.003$ for optimizing the reconstruction of the detailed surface boundary. We split the training/testing sets as in~\cite{nvf}.

\noindent\textbf{Metrics.} 
We adopt Chamfer Distance (CD), Normal Error (Normal), Earth Mover Distance (EMD), and F-score as evaluation metrics. We randomly sample 100k points with their corresponding normals from the reconstructed and ground-truth surfaces to compute CD ($\times 10^{-5}$) and Normal. EMD ($\times 10^{-2}$) is computed using 2048 randomly sampled points and F-score ($\times 10^{-2}$) uses 100,000 points with thresholds $1\times10^{-5}$ and $2\times10^{-5}$, respectively.


\subsection{Category-agnostic and Category-unseen Cases}
\label{sec:agnostic}
\noindent\textbf{Watertight shapes.}
We evaluate the model generalization ability of our NVF through category-agnostic and category-unseen reconstruction experiments. We report our quantitative experimental results on watertight shapes pre-processed by DISN~\cite{DISN} in~\cref{tab:watertight} to ensure a fair comparison as previous methods (\eg, OccNet~\cite{occnet} and IF-Net~\cite{ifnet}) cannot directly use non-watertight shapes as input. To maintain consistency with IF-Net~\cite{ifnet} and NVF~\cite{nvf}, we utilize 3k points as input for all methods and adopt the same training/testing split.

We provide quantitative qualitative comparison in~\cref{tab:watertight} and visualizations of the results in~\cref{fig:watertight}. 
Thanks to the richer cross-object prior information introduced from the codebook, our~\softname{} and~\hardname{} both outperform previous methods. Moreover, ~\softname{} has better model generalization ability for category-agnostic and category-unseen cases, therefore significantly surpassing UDF-based methods and ~\hardname{}. The qualitative analysis also demonstrates our method can produce smoother reconstructed surfaces than others due to curl regularization and direction loss. For example, our method can generate intricate details such as the tails of planes or concaves on speakers.

\noindent\textbf{Non-watertight shapes.} 
We also conduct additional experiments on the reconstruction of non-watertight shapes with the same experimental setup as described above. The results are presented in \cref{tab:non_watertight} and~\cref{fig:non_watertight}. We follow the same data splitting as in NVF~\cite{nvf}, where we randomly sample 3000 shapes from each base class for training and 200 shapes from each base and novel class for validation and testing. The input point cloud comprises 10k points sampled from the raw data of the shapes.

As shown in ~\cref{tab:non_watertight}, our methods ~\softname{} and ~\hardname{} outperform all existing methods regarding all matrices. In particular, ~\softname{} surpasses the most recent state-of-the-art method on all metrics. In particular, we achieve approximately $5\%$ and $15\%$ normal error reduction on base and novel classes compared with the previous best NDF and GIFS, respectively, which reflecting~\softname{} mitigate displacement inaccuracy and reconstruct smoother surfaces. The visualization in~\cref{fig:non_watertight} further reveals that NDF, our method produces a smoother reconstructed surface (~\eg, the center part for both $1^{st}$ and $2^{nd}$ row) with much fewer noises than other methods thanks to introduced curl and direction constraints.

\subsection{Category-specific Cases}
\label{sec:specific}
We present \textit{category-specific reconstruction} using ShapeNet cars, 
for demonstrating the efficacy in representing a single category. 
We follow the training/testing set and 10k-point sampling procedure as in~\cite{ndf,gifs,nvf}. 
As shown, on this task, our~\softname{} outperforms all other methods in comparison, including the second-best performer~\hardname{}, over four metrics, including CD, Normal, and two F-scores, while yielding comparable EMD to~\hardname{} and GIFS. Moreover, we demonstrate a qualitative visualization in~\cref{fig:car}, where our~\softname{} again produces smoother surfaces for finer parts than NDF, GIFS, and ~\hardname{}.

\subsection{Cross-domain Cases} 
\label{sec:cross}
In addition to the above intra-domain reconstruction tasks, we also examine our cross-domain ability. We devised a cross-domain reconstruction experiment and presented quantitative and qualitative results in~\cref{tab:garment} and~\cref{fig:garment}, respectively. Specifically, we assess the performance of our category-agnostic models, which were trained on ShapeNet base classes, on MGN~\cite{mgn} without fine-tuning. We randomly selected 20 shapes from the 5 categories in MGN and used 3,000 points from the raw data as input.

\cref{tab:garment} shows that our proposed~\softname{} outperforms other methods in terms of cross-domain reconstruction ability due to the better model generalization brought by a soft codebook than~\hardname{}. Specifically, our method significantly reduces approximately $10\%$ and $14\%$ on EMD, and $15\%$ and $45\%$ on normal metrics compared to~\hardname{} and GIFS, respectively. Moreover, the qualitative visualization in~\cref{fig:garment} shows that our approach accurately reconstructs the intricate folds and sharp edges of long coats as a better optimization by curl and direction losses, while other methods fail to capture.


\section{Analysis}

\noindent\textbf{Ablation.} 
We conduct comprehensive ablation studies on non-watertight category-agnostic and category-unseen reconstruction (same settings as~\cref{sec:agnostic}) to evaluate the effectiveness of the proposed components in~\cref{tab:abalation}, including hard/soft codebook (\ie, hard/soft), curl regularization (\ie, curl) and direction loss (\ie, dir). The results indicate that using codebooks can generally improve the overall performance regarding all metrics, and each introduced component can also improve the overall performance in most metrics.

\begin{table}[t]
\centering
\resizebox{0.5\textwidth}{!}{
    \begin{tabular}{@{}c|c|c|c|c|c|c@{}}
    \hline
     & Method & Backbone & CD$\downarrow$ & EMD$\downarrow$ & F1$_{1\times10^{-5}}\uparrow$ & F1$_{2\times10^{-5}}\uparrow$ \\
    \hline
     \multirow{5}{*}{Base} & NDF~\cite{ndf} & 3D Conv & 0.99 & 1.372 & 72.425 & 88.754 \\
     \cline{2-7}
     & \multirow{2}{*}{Ours (lite)} & 3D Conv & 0.92 & 1.189 & 72.415 & 89.365\\
      & & PointTransformer & 0.85 & 1.197 & 75.372 & 90.266 \\
     \cline{2-7}
     & \multirow{2}{*}{Ours (ultra)}& 3D Conv & 0.886 & \textbf{1.167} & 73.0827& 89.921 \\
      &  & PointTransformer & \textbf{0.810} & 1.182 & \textbf{76.256} & \textbf{90.785} \\
     \hline
     \multirow{5}{*}{Novel} & NDF~\cite{ndf} & 3D Conv & 0.93 & 1.532 & 76.162 & 89.977 \\
     \cline{2-7}
     & \multirow{2}{*}{Ours (lite)} & 3D Conv & 0.90 & 1.342 & 76.155 & 90.462\\
      & & PointTransformer & 0.78 & 1.340 & 79.723 & 91.576\\
      \cline{2-7}
     & \multirow{2}{*}{Ours (ultra)} & 3D Conv & 0.856 & \textbf{1.297} & 76.781 & 90.878\\
      &  & PointTransformer & \textbf{0.745} & 1.309 & \textbf{80.349} & \textbf{91.897}\\
     \hline
    \end{tabular}
    }
    \caption{Comparison of different backbones for Ours on non-watertight ShapeNet. The point cloud based backbones outperform the 3D convolution backbone.  PointTransformer performs the best. As the soft codebook and curl regularization require more memory, we only apply zero-curl regularization on the efficient backbone, \ie, PointTransformer.}
    \label{tab:backbone}
\end{table}



\noindent\textbf{Training Effectiveness.} We would like to highlight that the introduced ``hard'' codebook could also work as regularization to reduce the training time. As shown in~\cref{fig:loss}, the loss curves of the models with (w/) hard codebooks (\ie, blue line) generally converge faster than their corresponding models without (w/o) codebook (\ie, red line) or with (w/) soft codebook (\ie, green line). This observation could be attributed to the incorporation of the \textit{Exponential Moving Average} training strategy introduced by the ``hard" codebook, which serves as an additional momentum term and gradient regularizer during optimization. Such a mechanism does not present in the soft codebook and the vanilla version, potentially accounting for their relatively slower convergence than that of the hard codebook.

\noindent\textbf{3D Feature Extraction.} We also report our methods with different 3D feature extraction backbones (\ie, 3D convolution~\cite{ifnet,ndf,gifs} and PointTransformer~\cite{point_transformer}) in ~\cref{tab:backbone}.
The results show that our NVF can already achieve comparable (if not better) results with NDF using the same extraction backbone (\eg, 3D convolution), while ours is more efficient on runtime and memory cost. Using PointTransformer can consistently improve the overall performance. 

\begin{figure}[t]
  \centering
   \includegraphics[width=0.8\linewidth]{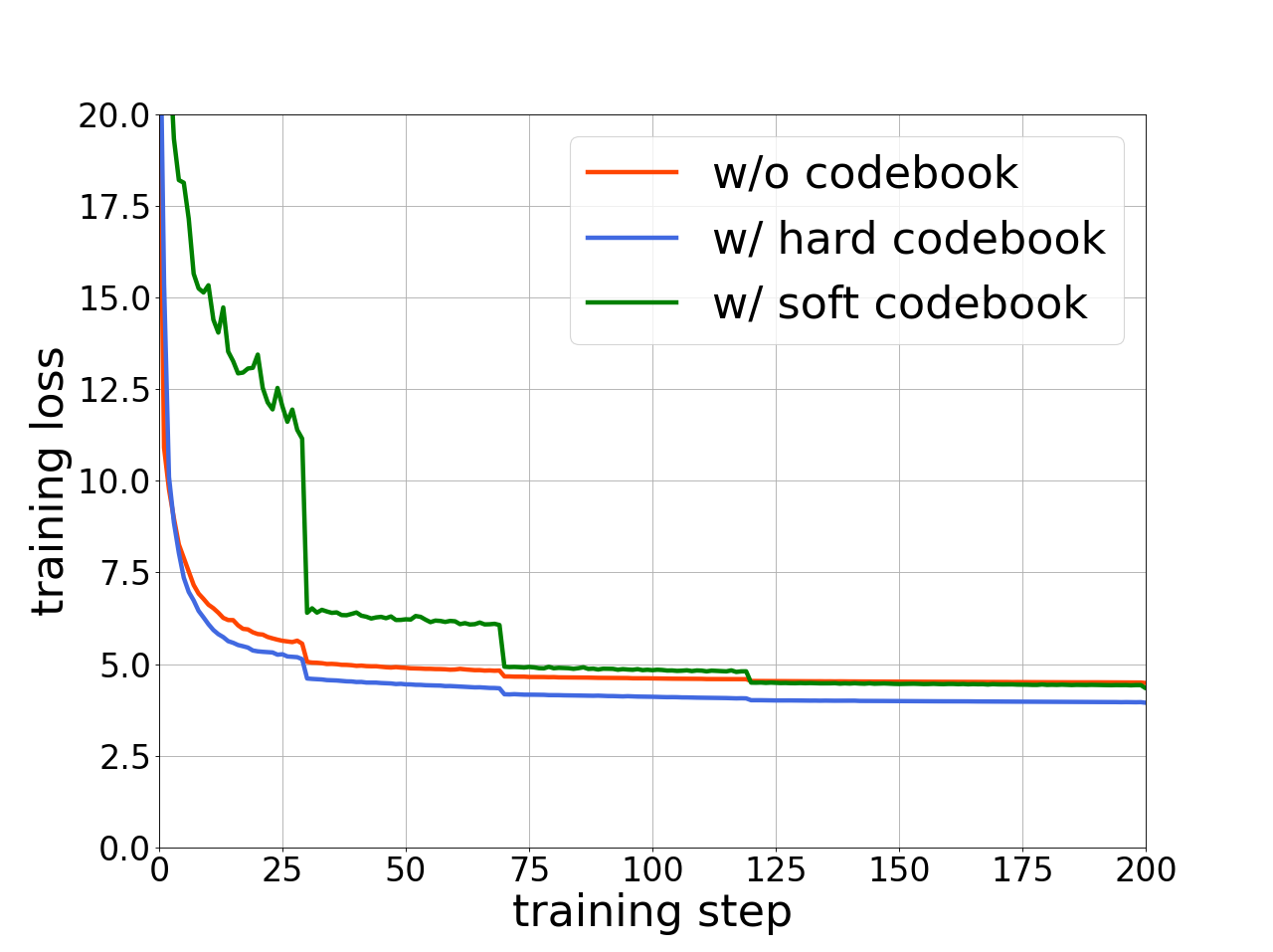}
   \caption{Training curves of models w/ and w/o codebooks. The models w/ hard codebooks converge faster than those w/o codebook or w/ soft codebook.}
   \label{fig:loss}
\end{figure}

\noindent\textbf{Soft vs Hard.} The benefits of the soft codebook over the conventional hard one can be seen in~\cref{tab:abalation} (with the same settings as~\cref{sec:agnostic}) on nearly all evaluation metrics for both the base and the novel classes, which is consistent with our analysis that the soft codebook enhances the representation ability. Nevertheless, it is important to note that this performance enhancement comes with a trade-off, as the utilization of the soft codebook incurs additional memory requirements compared to the hard one.


\noindent\textbf{Curl Loss.} We set a small weight of the curl regularization to balance the magnitudes of different loss terms, and showed two visual examples in the~\cref{fig:rotation}, demonstrating reconstruction results w/o (\cref{fig:rotation}\textcolor{red}{(c)}) and w/ (\cref{fig:rotation}\textcolor{red}{(d)}) using curl regularization. Quantitatively, on non-watertight test objects in ShapeNet, our model produces curl values of 1.256 and 1.404 w/ and w/o using curl regularization for optimization, respectively.

\begin{table}[t]
\centering
\resizebox{0.4\textwidth}{!}{
    \begin{tabular}{@{}c|c|c|c@{}}
    \hline
    Methods & Backbone & Runtime & Memory \\
    \hline
    NDF~\cite{ndf}& 3D Conv & 0.327s & 7.30G \\
    \hline
    Ours (lite) & PointTransformer& 0.113s & 2.35G \\
    Ours (ultra) & PointTransformer & 0.115s & 15.48G \\
     \hline
    \end{tabular}
    }
    \caption{Inference analysis. The runtime and memory are time cost and peak memory during the inference of 50k queries.}
    \label{tab:speed}
\end{table}

\noindent\textbf{Complexity.} We report the inference time and peak GPU memory on the machine with one RTX 3090Ti in~\cref{tab:speed} for obtaining the distance and direction of 50k query points. The results show that our~\softname{} requires $40\%$ inference time compared with NDF, and an equal level of inference time compared with~\hardname{}, while achieving better performance. However, our~\softname{} needs more memory compared with NDF and~\hardname{}. The underlying reason for this predicament is attributed to the memory overhead associated with the computation of attention maps in the soft codebook. To address this challenge, one possible solution is to adopt a sparse attention mechanism, which would greatly alleviate this computational burden. We aim to investigate and develop such an approach in our forthcoming research endeavors.

\begin{figure}[t]
  \centering
   \includegraphics[width=0.95\linewidth]{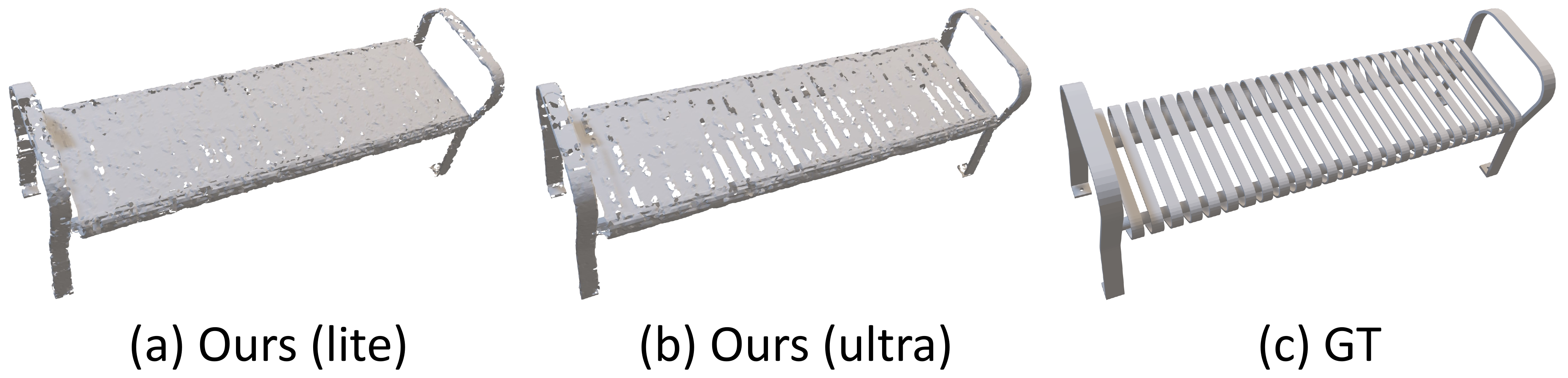}
   \vspace{-1mm}
   \caption{Visualization of our failure case example, when reconstructing a very thin, complex, and high-frequency 3D structure.}
   \label{fig:failure}
\end{figure}

\noindent\textbf{Limitations.} It is noteworthy that our proposed methodology has the capacity to effectively manage shapes with arbitrary resolution and topology, regardless of whether they are watertight or non-watertight. It is important to acknowledge, however, that there are still certain limitations associated with the reconstruction process, particularly when it comes to very thin, high-frequency, and intricate structures. As evidenced by the example of the bench beam in~\cref{fig:failure}, there are still areas where the methodology falls short of achieving optimal outcomes.

\begin{figure}[t]
  \centering
  \includegraphics[width=0.95\linewidth]{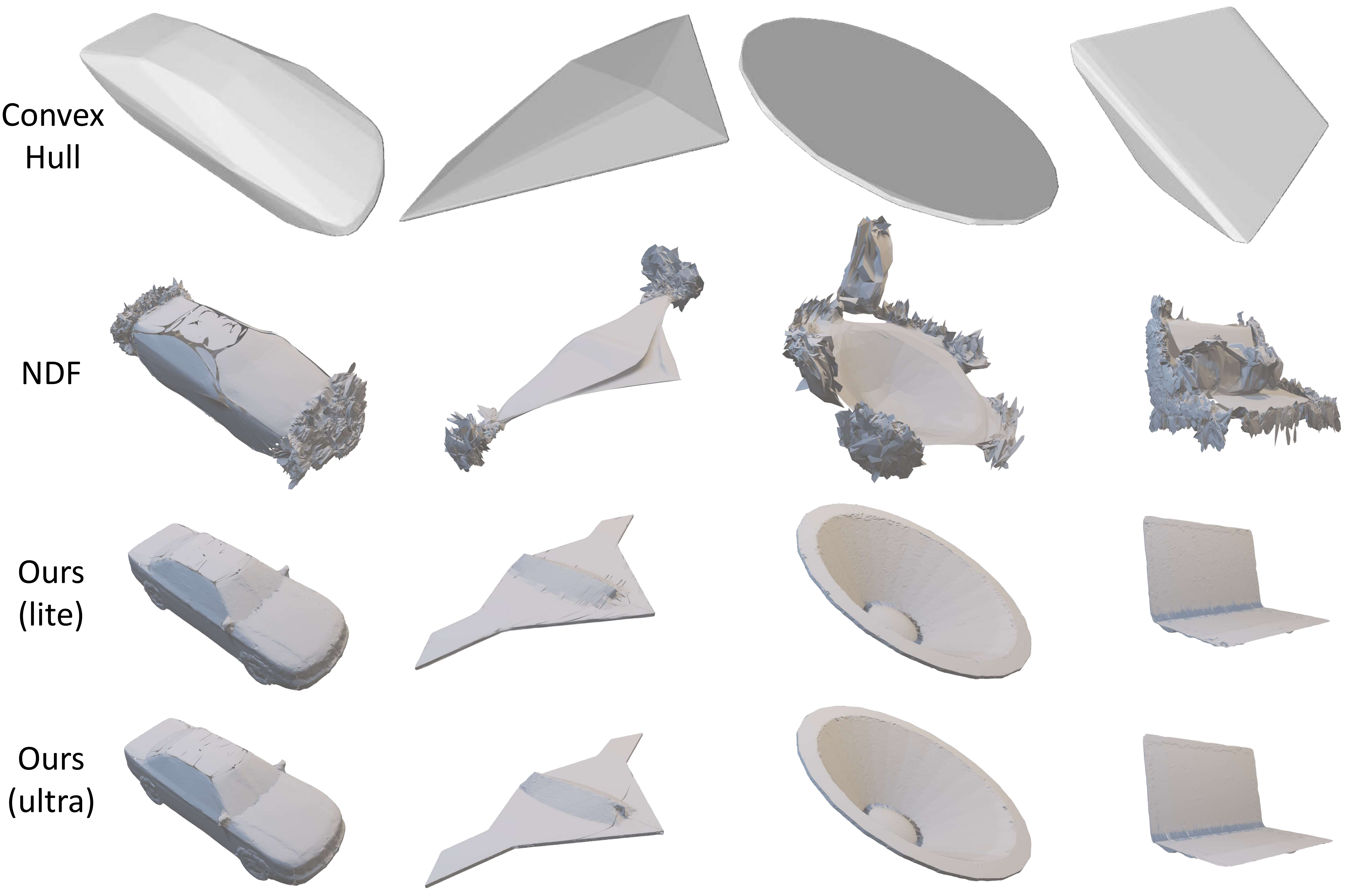}
  \caption{Mesh deformation visualization. The $1^{st}$ row represents the source meshes that are the convex hull of the input point clouds, and the $2^{nd}$ row represents the shapes after deformation by our NVF.}
   \label{fig:deform}
\end{figure}
\noindent\textbf{Explicit Deformation Visualization.} In order to demonstrate that NVF could serve as both an implicit neural function and an explicit deformation function (\ie, displacement output of the function could be directly used to deform source meshes), we provide some deformation results in~\cref{fig:deform}, which indicates that our NVF can directly deform the convex hull meshes of the input point cloud to fit the surface.

\section{Conclusion}

In this paper, we present a novel framework for surface reconstruction representation, named Neural Vector Fields (NVFs). Our approach employs hard or soft codebooks to introduce cross-object priors, enriches the feature representation for reconstruction, and enables the differentiability of the framework. Moreover, we propose that the ideal Distance Vector Fields (DVFs) should be active and irrotational fields, and we observe that inaccurate displacement predictions lead to uneven surfaces and little holes in reconstructed surfaces. To address this issue, we inject curl regularization and direction loss to better supervise the learning process, leveraging the differentiable soft codebook design which allows for Jacobian matrix calculation. Experimental results demonstrate that our method achieves state-of-the-art reconstruction performance for various topologies.

{\small
\bibliographystyle{ieee_fullname}
\bibliography{egbib}
}

\end{document}